# Color Image Classification via Quaternion Principal Component Analysis Network

Rui Zeng[1, 4], Jiasong Wu [1, 2, 3, 4], Zhuhong Shao[1, 4], Yang Chen[1, 4], Lotfi Senhadji[2, 3, 4],

Huazhong Shu[1, 4]

[1]*LIST, the Key Laboratory of Computer Network and Information Integration, Southeast University, Ministry of Education, Nanjing 210096, China*

[2]*INSERM, U 1099, Rennes 35000, France*

[3]*Laboratoire Traitement du Signal et de l'Image, Université de Rennes 1, Rennes 35000, France*

[4]*Centre de Recherche en Information Biomédicale Sino-français, Nanjing 210096, China*

**Abstract**: The Principal Component Analysis Network (PCANet), which is one of the recently proposed deep learning architectures, achieves the state-of-the-art classification accuracy in various databases. However, the performance of PCANet may be degraded when dealing with color images. In this paper, a Quaternion Principal Component Analysis Network (QPCANet), which is an extension of PCANet, is proposed for color images classification. Compared to PCANet, the proposed QPCANet takes into account the spatial distribution information of color images and ensures larger amount of intra-class invariance of color images. Experiments conducted on different color image datasets such as Caltech-101, UC Merced Land Use, Georgia Tech face and CURet have revealed that the proposed QPCANet achieves higher classification accuracy than PCANet.

## 1. Introduction

Image classification is among the very active research topics in the field of pattern recognition and computer vision. It is a challenging task because image contents are subject to various changes in terms of illumination, rotation, scaling or more complex deformation. To effectively solve this problem, numerous approaches have been proposed in the past decades. They can be divided into two groups: the first one consists of extracting manually low-level intrinsic features, and the second one relies on deep architectures for learning from data sets of interest. For the first group, the most commonly used methods are based on scale-invariant feature transforms [1], [2]. The second group, which is considered as a reasonable way to overcome the limitations of manual extraction of features, has became an important research focus in recent years. The objective of deep learning [3] is to extract intrinsic features from data by multi-level architecture, with the hope that higher-level features represent more concise semantics of the data. There are several approaches to build deep architectures such as deep belief networks (DBN), auto-encoders (AE), etc. DBN [4] is a multilayer generative model, which uses an unsupervised learning algorithm and greedily trains one layer at a time. After this learning step, it is trained to perform classification in a supervised way. Hinton and Zemel [5] proposed the AE approach, which is trained locally by means of an unsupervised learning algorithm.

The key factor of the success of deep learning in image classification is the use of convolutional architecture [6-12]. Convolutional deep neural network was inspired by the structure of the visual system. An example of such an architecture is the convolutional neural networks (CNN) [7, 13] using the error gradient. It allows achieving the state-of-the-art performance on several pattern recognition tasks. LeCun's CNN consists of multiple trainable stages stacked on top of each other, followed by a supervised classifier. Each stage is organized in layers of two types: convolutional layers and subsampling layers. Recently, Mallat and Bruna [14, 15] proposed a wavelet scattering network (ScatNet) built upon prefixed wavelets as convolutional filters and modulus transform as a nonlinear operators. Hence there is no need for learning parameters at all. Unlike other empirically justified deep learning algorithms, ScatNet [14, 15] is mathematically justified. More recently, Chan et al. [16] proposed a new deep learning architecture, namely, Principal Components Analysis Network (PCANet). It uses the most basic and simple operations to emulate the processing layers of CNN. In fact, the data-adapting filter bank of convolutional layers in each stage is obtained by the principal components analysis (PCA) filters. The nonlinear operation used here is the binary hashing. In the last stage of PCANet, the block-wise histograms of the binary codes are used for pooling. Surprisingly, such a simple deep architecture is quite on par with and often better than the state-of-the-art techniques of features selection (prefixed, hand-crafted, or learned from deep architecture) for most image classification tasks, including face images, hand-written digits, texture images, and object images.

Many researchers [17-19] pointed out that color images can provide a large amount of information of the real-world objects. However, most of the above classification algorithms (CNN, ScatNet, PCANet, etc.) were only proposed for gray images. Thus, similar researches that are suitable for color images classification need to be conducted.

The classical representation of color images usually combines the values of R, G, B - channels into one vector. Under this simple representation, the spatial relationships between the color pixels of the image are destroyed, and the dimension of images is three times of that of classical gray-scale images. As a result, it is crucial to seek a way to better represent color images by taking into consideration the spatial relationships between R, G, B channels. Quaternion is a powerful mathematical tool that allows representing color images [20-23]. In this way, color images can be represented by quaternion as a whole and the intrinsic structures of color images and the relationships between R, G, B channels can thus be preserved.

On the other hand, principal components analysis concepts have been extended to quaternion algebra in recent years [24, 25]. Bihan and Sangwine [24] and Pei et al. [25] proposed a new feature extraction method for color image, called Quaternion Principal Components Analysis (QPCA), which is able to extract more robust and informative features from color image than classical PCA. These studies emphasize that QPCA is superior to PCA in color image representation.

In this paper, we propose a new color image feature extraction technique, namely Quaternion Principal Component Analysis Network (QPCANet) that extends and adapts the principals of PCANet approach to quaternions. We represent color images in quaternion domain and extract QPCA features, which are then put into simple quaternion binary hashing to get final QPCANet features. Support Vector Machine (SVM) [26, 27] is finally used as feature classifier. The performances of the proposed method (i.e QPCANet) are then evaluated and compared to those of PCANet on color image databases for many classification tasks, including face recognition, object recognition, texture classification and land use classification.

The rest of the paper is organized as follows. In Section 2, the quaternion algebra is briefly introduced as well as its use for color image representation. Section 3 is dedicated to the presentation of the architecture of QPCANet and also shows how the multi-stage QPCANet works. In Section 4, classification performances of QPCANet and PCANet are evaluated and compared on various color datasets. The paper ends with concluding remarks reported in Section 5.

## 2. Quaternion algebra and quaternion representation of color image

The field of quaternion numbers is denoted by $\mathbb{H}$. A quaternion number $x$ is a hypercomplex number, which consists of one real part and three imaginary parts:

$$x = S(x) + I(x)i + J(x)j + K(x)k \tag{1}$$

where $S(x), I(x), J(x), K(x) \in \mathbb{R}$ ($\mathbb{R}$ denotes the field of real numbers) and $i$, $j$ and $k$ are three imaginary units obeying the following rules:

$$i^2 = j^2 = k^2 = ijk = -1, ij = -ji = k, jk = -kj = i, ki = -ik = j \tag{2}$$

Eq. (1) can also be expressed using the following 2-tuple or 4-tuple notations:

$$x = (S(x) \quad \mathbf{v}) = (S(x) \quad I(x) \quad J(x) \quad K(x)) \tag{3}$$

where $S(x)$ is a scalar part of $x$ and $\mathbf{v}$ forms the vector part. For the special case where $S(x) = 0$, i.e. $x = I(x)i + J(x)j + K(x)k$, $x$ is called a pure quaternion.

The conjugate of a quaternion is defined as:

$$x^* = (S(x) \quad -\mathbf{v}) = (S(x) \quad -I(x) \quad -J(x) \quad -K(x)) \tag{4}$$

The $l_2$ norm of a quaternion $x$ is defined as $\|x\| = \sqrt{S^2(x) + I^2(x) + J^2(x) + K^2(x)}$ and its inverse is given by $x^{-1} = \dfrac{x^*}{\|x\|^2}$. If $\|x\| = 1$, then $x$ is called a unit quaternion. An important characteristic of the quaternion algebra is its non-commutatively under multiplication. For a complete review of quaternion properties, please refer to [28].

A color image can be represented in the quaternion domain as follow [20]:

$$Q(y, z) = R(y, z)i + G(y, z)j + B(y, z)k \tag{5}$$

where $R(y, z)$, $G(y, z)$ and $B(y, z)$ are respectively the red, green, and blue components of the pixel at position $(y, z)$. Thus each color pixel is represented as a pure quaternion and its imaginary parts represent the color channels. This gives birth to the quaternion image Q.

## 3. Quaternion Principal Component Analysis Network (QPCANet)

The architecture of the proposed QPCANet is depicted in Fig. 1. In this section, we analyze the structure of QPCANet for color image classification, and we show how to build a multi-stage QPCANet.

### 3.1 QPCA filter bank

Suppose that we have $N$ quaternion input patterns $\{\mathbf{Q}_i\}_{i=1}^{N}$ of size $m \times n$ and corresponding labels for training. Quaternion input pattern can either be a quaternion image or a meaningful quaternion matrix. For simplicity, we assume that the patch size is $k_1 \times k_2$, $k_1$ and $k_2$ are positive odd integers. We collect all $(m-k_1+1) \times (n-k_2+1)$ quaternion patches around each pixel of the *i*th quaternion input pattern. Then each quaternion patch is centered by subtracting its mean. We get thus zero-mean quaternion patches for the *i*th quaternion input pattern. We note these quaternion features $\mathbf{q}_i = [\mathbf{q}_{i,1}, \mathbf{q}_{i,2}, \ldots, \mathbf{q}_{i,(m-k_1+1)\times(n-k_2+1)}]$, and each element of $\mathbf{q}_i$ belongs to $\mathbb{H}^{k_1 k_2}$. Repeating the above process, we can get all quaternion patches of $N$ input patterns for training. By constructing the same matrix for all quaternion patches and by putting them together, we obtain

$$\mathbf{q} = [\mathbf{q}_1, \mathbf{q}_2, \ldots, \mathbf{q}_N] \in \mathbb{H}^{k_1 k_2 \times N(m-k_1+1)(n-k_2+1)} \tag{6}$$

The covariance matrix of $\mathbf{q}$ is computed as:

$$\Sigma = \frac{\mathbf{q}\mathbf{q}^*}{N(m-k_1+1)(n-k_2+1)} \tag{7}$$

The above matrix admits a quaternion eigenvalue decomposition [24, 25, 29, 30]:

$$\Sigma = \mathbf{W}\Omega\mathbf{W}^* \tag{8}$$

where $\mathbf{W} \in \mathbb{H}^{k_1 k_2 \times k_1 k_2}$ is a unitary matrix that contains the eigenvectors of $\Sigma$ and $\Omega \in \square^{k_1 k_2 \times k_1 k_2}$ is a diagonal matrix with eigenvalues on its diagonal. The values on the diagonal of $\Omega$ are arranged in decreasing magnitude order and the corresponding eigenvectors, i.e. the principal components of $\mathbf{q}$, are arranged accordingly in $\mathbf{W}$. The bigger the eigenvalue is, the more important the quaternion principal component will be. These principal component vectors are also called QPCA filters or QPCA filter bank. It was shown in [24] that QPCA is invariant to spatial rotation of color image.

Let $L$ be the desired number of QPCA filters. We choose then the first $L$ eigenvectors of $\mathbf{W}$ to form the new matrix $\bar{\mathbf{W}} \in \mathbb{H}^{k_1 k_2 \times L}$ where each of its columns is seen as a filter $\mathbf{W}_l$:

$$\mathbf{W}_l \in \mathbb{H}^{k_1 \times k_2}, l = 1, 2, \ldots, L \tag{9}$$

These filters form the filter bank of quaternion input patterns $\{\mathbf{Q}_i\}_{i=1}^N$. This filter bank captures the main variations of all quaternion patches. We will use the filter bank to extract the feature maps from quaternion input patterns by convolutional operations.

*3.2 Convolutional operation*

A quaternion input pattern $\mathbf{Q}_i$ is convolved with the above filter bank to form quaternion feature maps:

$$\mathbf{F}_i^l = \mathbf{Q}_i * \mathbf{W}_l, i = 1, 2, \ldots, N, l = 1, 2, \ldots, L \tag{10}$$

where $*$ denotes 2D quaternion convolution. Note that the 2D convolution of quaternion matrices is also linear. The boundary of $\mathbf{Q}_i$ is padded by zero quaternion (i.e. zero-padding), whose real part and three imaginary parts are zeros, so as to ensure that $\mathbf{F}_i^l$ and $\mathbf{Q}_i$ have the size $m \times n$. $\mathbf{Q}_i$ can then be transformed into several quaternion feature maps $\mathbf{F}_i^l$ according to the numbers of QPCA filters. We denote by $\{\mathbf{F}_i\} = \{\mathbf{F}_i^l\}_{l=1}^{L}$ the set of quaternion feature maps of $\mathbf{Q}_i$. It turns out that each element of $\{\mathbf{F}_i\}_{i=1}^{N}$ can also be used as quaternion input pattern. By repeating the above process one can expect to derive high-level features.

*3.3. Quaternion feature maps weighting and pooling*

Quaternion feature maps extracted by convolution operation should be binarized, weighted, and then summed in order to reduce the number of quaternion feature maps. Thus, each quaternion feature map $\mathbf{F}_i^l = S(\mathbf{F}_i^l) + I(\mathbf{F}_i^l)i + J(\mathbf{F}_i^l)j + K(\mathbf{F}_i^l)k$, is binarized by applying the Heaviside step function $H(\cdot)$ to its four parts (the value of $H(\cdot)$ is one for positive entries and zero otherwise). The binarized quaternion feature map is denoted by $\bar{\mathbf{F}}_i^l = S(\bar{\mathbf{F}}_i^l) + I(\bar{\mathbf{F}}_i^l)i + J(\bar{\mathbf{F}}_i^l)j + K(\bar{\mathbf{F}}_i^l)k$ and the resulting maps are weighted to form a new single quaternion pattern:

$$\mathbf{T_i} = \sum_{l=1}^{L} 2^{l-1} \bar{\mathbf{F}}_i^l = S(\mathbf{T}_i) + I(\mathbf{T}_i)i + J(\mathbf{T}_i)j + K(\mathbf{T}_i)k. \tag{11}$$

Note that the pixel values of $S(\mathbf{T}_i), I(\mathbf{T}_i), J(\mathbf{T}_i), K(\mathbf{T}_i)$ are integers belonging to the interval

$[0, 2^L -1]$.

Next, we discuss the pooling stage of QPCANet. Since $\mathbf{T}_i$ has four parts, namely, $S(\mathbf{T}_i), I(\mathbf{T}_i), J(\mathbf{T}_i), K(\mathbf{T}_i)$, each part is pooled separately. Let us first consider $S(\mathbf{T}_i)$ for example. We divide $S(\mathbf{T}_i)$ into $B$ blocks. For each block, we perform a pooling operation, which is simply the estimation of the histogram (with $2^L -1$ bins) of the decimal values in each block. After this pooling process, we concatenate the histograms of the $B$ blocks into one vector, that is, $B\text{hist}(S(\mathbf{T}_i)) \in \mathbb{R}^{2^L B}$. After performing the same pooling operation on $I(\mathbf{T}_i), J(\mathbf{T}_i), K(\mathbf{T}_i)$, the obtained four vectors are concatenated to derive the feature vector, of the quaternion input pattern $\mathbf{Q}_i$, denoted by:

$$\mathbf{f}_{1i} = [B\text{hist}(S(\mathbf{T}_i)), B\text{hist}(I(\mathbf{T}_i)), B\text{hist}(J(\mathbf{T}_i)), B\text{hist}(K(\mathbf{T}_i))] \in \mathbb{R}^{4 \times 2^L B} \quad (12)$$

Chan et al. [16] showed that, depending on applications, the local blocks can be either overlapping or non-overlapping in PCANet. They suggested that non-overlapping blocks are suitable for face images, whereas the overlapping blocks are appropriate for hand-written digits, textures, and object images. We found that the recommendation is also suitable for QPCANet.

*3.4. Multi-stage architecture*

Now we are ready to describe the multi-stage architecture of QPCANet. As depicted in Fig. 1, the two-stage QPCANet (QPCANet-2) contains two QPCA filter banks ($\mathbf{W}^1$ and $\mathbf{W}^2$), two convolution layers (C1 and C2), a quaternion feature map weighting layer and pooling layer. The output of the last layer is fed to SVM [26, 27] for classification. The QPCA filter bank $\mathbf{W}^1$, which contains $L_1$ QPCA filters, can be obtained from $\{\mathbf{Q}_i\}_{i=1}^N$, then layer C1 executes a convolution using the kernel $\mathbf{W}^1$ to get the sets of quaternion feature maps $\{\mathbf{F}_i\}_{i=1}^N$. The filter bank $\mathbf{W}^2$, which contains $L_2$ QPCA filters, is computed from $\{\mathbf{F}_i\}_{i=1}^N$. The layer C2 executes a convolution using $\mathbf{W}^2$ to get $L_1 \times L_2$ quaternion maps $\mathbf{G}_i^{11}, \ldots, \mathbf{G}_i^{1L_2}, \mathbf{G}_i^{21}, \ldots, \mathbf{G}_i^{2L_2}, \mathbf{G}_i^{L_1 1}, \ldots, \mathbf{G}_i^{L_1 L_2} \in \mathbb{H}^{m \times n}$, where $\mathbf{G}_i^{ab}$ denotes the $b$th quaternion feature map of $F_i^a$ for $\mathbf{Q}_i$. Each $\mathbf{G}_i^{ab}$ is binarized and weighted as previously described. We then pool the $L_1 \times L_2$ quaternion feature maps, using the method reported in Section 3.3, to obtain the final feature vector $\mathbf{f}_{2i} \in \mathbb{R}^{4 \times 2^{L_2} L_1 B}$.

One or more additional stages can be stacked like $\mathbf{W}^1$-C1-$\mathbf{W}^2$-C2-$\mathbf{W}^3$-C3…. But in our experiments, we found that QPCANet-2 is enough to get a good accuracy in color image classification and thus a deep architecture is not necessarily required.

## 4. Experimental results

To evaluate the efficiency and robustness of the proposed scheme, a set of experiments has been conducted on various datasets (Caltech-101, UC Merced Land Use, Georgia Tech face database and CURet texture database).

Chan et al. [16] compared PCANet with many other methods (CNN [7], ScatNet [14, 15], etc) for the classification of gray-scale images. Therefore, in this study, we only compare the proposed QPCANet with PCANet for color image classification. Note that when using the PCANet to deal with color image, both the color-to-gray transformation (Gray PCANet) and three channels concatenation representation (RGB PCANet) are taken into consideration.

### 4.1. Performance in Caltech-101

We evaluate the performance of QPCANet on Caltech-101 dataset [31] for color image recognition. Caltech-101 contains a total of 9146 images including color and gray-level images, belonging to 101 distinct objects, including faces, watches, ants, pianos, etc. In this paper, we only consider 8733 color images in Caltech-101 dataset and discard all gray-level images. All color images were resized to be $32 \times 32$ pixels in order to reduce the computational time. We randomly select 15 color images as training images per class and the remaining ones are used for testing. The overlapping ratio of block is set to 0.5 to preserve appropriate feature invariance for objects in Caltech-101. The block size in feature pooling stage is set to be $7 \times 7$, based on the experience [16]. Then we use a greedy algorithm to find the optimal number of QPCA filters in every stage of QPCANet by changing the patch size from $3 \times 3$ to $7 \times 7$. First we vary the $L_1$ from 2 to 9 for QPCANet-1. The results are shown in Fig. 2(a). One can see that QPCANet-1 achieves the best result for patch size $3 \times 3$ and the accuracy of $L_1 = 8$ is satisfied. Although the accuracy increases with the increase of $L_1$ for other patch size, the optimal number of QPCANet-1 is set to 8, which is the same as that of PCANet.

Next, two-staged QPCANet is considered. We set $L_1 = 8$ and vary $L_2$ from 2 to 9. The patch size is changed from $3 \times 3$ to $7 \times 7$. We draw the accuracy of QPCANet-2 in Fig. 2(b). It can be seen that QPCANet achieves a good performance for $L_2 = 8$. When $L_2 \geq 8$, the recognition rate for patch size $3 \times 3$ and $7 \times 7$ decreases, and there is no obvious increase in accuracy for patch size $5 \times 5$. Considering the applications of QPCANet, we should make the parameters in this system as simple as possible. We have noticed empirically that even with simple parameters, the system can achieve good performance in color image classification. So, although some fine-tuned $L_1$ and $L_2$ values could lead to performance improvement, we decided to set $L_1 = L_2$.

For fair comparisons, appropriate parameters of RGB PCANet and gray PCANet are required. For both RGB PCANet and Gray PCANet, we enlarged patch size from $3 \times 3$ to $7 \times 7$. The block size was fixed to $7 \times 7$ and overlapping ratio is set to 0.5. The feature maps extracted from the above networks were used with SVM [26, 27] as a classifier. The average accuracy over 5 drawing of the training set is listed in Table 1. We see that the best performance (63.78%) is achieved by QPCANet-2 when patch size is $3 \times 3$. With the increase of the patch size the accuracy decreases. This illustrates that too big patch size is not appropriate for color object recognition with QPCANet. The performance improvement from QPCANet-1 to QPCANet-2 is not as large as that of RGB PCANet. But we note that the accuracy of QPCANet-1 is much better than other methods. IN fact, the behavior of QPCANet-1 is very close to that of Gray PCANet-2 and RGB PCANet-2. This implies that we can achieve an acceptable result by just apply a simple one-stage network of QPCA.

By comparing Gray PCANet with RGB PCANet, a strange phenomenon draws our attention. Why the accuracy of RGB PCANet-1 is lower than that of Gray PCANet-1, but RGB PCANet-2 outperforms Gray PCANet-2? First, let us consider some pictures extracted from Caltech-101 and reported in Fig. 3. We note that every horizontal pair of objects has the same main colors, but the objects are not belonging the same class. Thus, we cannot do a right classification only with color information. For color image classification, color information and boundary information seem equally important. Color information is harmful to color image classification when an algorithm has not enough representation ability. That is, color is a burden for some classification algorithms. It seems that RGB PCANet-1 is such an algorithm, which is not able to utilize simultaneously color information and boundary information. For Gray PCANet-1, we changed color images to gray-scale ones and thus only the boundary information, which is effectively utilized by Gray PCANet-1, is preserved. Therefore, it is not surprising that Gray PCANet-1 outperforms RGB PCANet-1.

The performance improvement from RGB PCANet-1 to RGB PCANet-2 is surprising. RGB PCANet-2 seems to have enough ability to deal with both color information and boundary information. Compared to RGB PCANet-1, the representation ability of RGB PCANet-2 is significantly higher. RGB PCANet-2 also performs better than Gray PCANet-2 by utilizing color information of objects. Gray PCANet-2 does not perform significantly better than Gray PCANet-1 in this database. We can conclude that the accuracy of color object recognition is difficult to improve by only utilizing boundary information.

Why QPCANet outperforms the two other methods (RGB PCANet, Gray PCANet) in color image classification? We think that the reason is the utilization of quaternion model of color image shown in Eq. (5). The quaternion representation of color image preserves the underlying spatial structures and relationships between R, G, B channels. QPCA is shown to be more suitable for color images representation, and allows enhancing the robustness of color images features [29]. The shortcoming of RGB PCANet is that it neglects the structure and unity of color images.

To conceptualize the learned QPCANet filters of patch size $3 \times 3$, we draw them in Fig. 4. In this figure, the first four rows correspond to the real part and three imaginary parts of the first stage QPCA filters, and the others correspond to the four parts of the second stage QPCA filters. The learned RGB PCANet filters of patch size $3 \times 3$ are shown in Fig. 5. The first three rows represent the one stage RGB PCA filters where each row corresponds to R, G, B channels, respectively. The last row represents the PCA filters of the second stage.

All the filters depicted in figure 4 and 5 are scaled within [0, 1]. Some learned filters exhibit similar shapes however; their numerical values are not the same. We notice that some similar QPCA filters occur several times in the different parts of quaternion. Such a filter redundancies may improve the intra-class invariance of feature maps. For the first three rows in RGB PCA filters, we notice that the last seven RGB PCA filters of each row have the same shape although their numerical values are not the same. It means that the similar filtering operations are performed in R, G, B channels.

*4.2. Testing on UC Merced land use database*

Subsequently, we test the proposed QPCANet performance on UC Merced land use dataset [32]. UC Merced land use dataset is manually extracted from large images from the USGS National Map Urban Area Imagery collection for various urban areas around the country. The dataset, composed of $256 \times 256$ pixels RGB images with pixel resolution of one foot, was and manually partitioned into 21 classes. Each class contains 100 images. Fig. 6 shows 21 images each representing a distinct class. Some representative images from the dataset are also shown in Fig. 7, from which we find that the dataset contains abundant rotation information due to a different point of view of aerial photography. The land objects in one class are the mainly the same, but the angles of view are different. The intra-class variability will then increase the difficulty of color image classification.

The QPCANet is trained with the number of filters $L_1 = L_2 = 8$, block size is set to $8 \times 8$ and the overlapping ratio is fixed to 0.6. We changed patch size $k_1 \times k_2$ from $3 \times 3$ to $7 \times 7$. Every image was scaled to $32 \times 32$ pixels to reduce computational time. We randomly selected 80 training images per class, and remaining ones were used for testing. We used the same parameters for RGB PCANet-2 and Gray PCANet. The results are reported in Table 2. Note that with the increase of patch size, the accuracy of all networks decreases. The best performance for various patch sizes is highlighted in bold. In UC Merced Land use dataset, one stage QPCANet outperforms all other one-stage networks. It is even better than some two-stage networks. This implies that only one-stage QPCANet can achieve acceptable results. QPCANet-2 for patch size $3 \times 3$ helps increasing the performances by more than 6% of those of RGB PCANet-2. The Gray PCANet performs poorly in all cases.

It turns out that QPCANet performs much better than both RGB PCANet and Gray PCANet in UC Merced Land Use database. The reason may be due to the nature of UC Merced land use database with contains a large amount of rotation type images. Quaternion representation gives a concise and all-in-one representation for rotation, and is known for its successful application in computer graphics, computer vision, and orbital mechanics of satellites, etc. [33-37]. We suspect that the quaternion model of color image gives more rotation invariance to QPCANet when compared to RGB PCANet. In the next section, we further experimentally verify this conjecture.

*4.3. Face recognition on Georgia Tech face database*

Georgia Tech face database [38] contains images of 50 people in JPEG format. For each individual, 15 color images were collected at the Center for Signal and Image Processing at Georgia Institute of Technology. All images in the database are $640 \times 480$ pixels and the average size of faces in these images is $150 \times 150$ pixels. Most of the images were taken in two different sessions to take into account the variations in illumination conditions, facial expression, and appearance.

In this experiment, the face images are cropped with dimension 128 × 160 to guarantee classification accuracy. Some examples are shown in Fig. 8. To reduce the computational complexity, we resized all cropped images to 64 × 64 pixels. We randomly pick up 10 images in each class as train images, and the others are used as testing images. We found that one-stage networks provide excellent results. Therefore, two-stage networks are not considered here. The number of filters in networks is set to 8 and non-overlapping block is of size 8 × 8. The average accuracy of classifications over 10 times experiments are listed in Table 3.

Next, we experimentally verify the hypotheses in UC Merced Land use: QPCANet outperforms RGB PCANet when color images to classify are related to each other by rotation.

First, we choose only one cropped front face image from 20 individuals. These images are zero padded to form a square image of size 220 × 220 pixels. A representative padded image is shown in Fig. 9 (a). The padding ensures that the rotation operation will not induce any border effect. For each individual (i.e. class), we rotate the padded front image from 0° to 360° by a step of 10° to obtain 36 images for each class. Some rotated images are shown in Fig. 9 (b). We randomly select 18 images from each class to train networks and the remaining ones are used in the testing procedure. All images are linearly scaled to 64 × 64 pixels. For all networks, the number of filters, and the (non-overlapping) block size are set to $L_1 = L_2 = 8$, and 8×8, respectively. The results of QPCANet and other networks are listed in Table 4. Surprisingly, the accuracy of all networks decreases with the increase of stage. When patch size is 5 × 5, the accuracy of QPCANet-1 reaches 84.94% outperforming all the other methods. We can conclude that, compared to PCANet, QPCANet allows extracting features that are less sensitive to rotation. .

*4.4. Texture discrimination in CURet Dataset*

The CURet texture dataset [39] contains images of 61 materials that broadly span the range of different surfaces that we commonly see in our environment. Every class is composed of 92 images of size 200 × 200 pixels. All images are linearly scaled to 32 × 32 pixels. The database is randomly splited into a training set and a testing set, with 23 training images for each class.

The QPCANet is trained with a number of filters $L_1 = L_2 = 8$, the overlapping ratio is set to 0.5 and the block size is fixed $8 \times 8$. The RGB PCANet and Gray PCANet use the same parameters as that of QPCANet.

The average recognition rates over 5 different random splits are given in Table 5. We see that QPCANet outperforms other networks whether the stage is one or two. When patch size is set to $7 \times 7$, the performance of QPCANet-1 approximately equals that of RGB PCANet-2 and outperform that of Gray PCANet-2. We also notice that RGB PCANet-1 outperforms Gray PCANet-1. These performances are not the same as objects recognition in Caltech-101 because texture dataset does not have such complex property as object dataset. The performance of RGB PCANet is much higher than that of Gray PCANet due to the use of color information and boundary information of texture dataset at the same time.

## 5. Conclusion

In this paper, we have proposed QPCANet, a simple and tractable model for color image classification. QPCANet relies upon three steps: QPCA filter bank, convolutional operation, quaternion feature maps weighting and pooling. In QPCA filter bank, quaternion component are extracted from quaternion representation of color images. In convolution operation, the quaternion components extracted from QPCA filter bank are used to form several quaternion feature maps. In the quaternion features weighting and pooling step, quaternion features are binarized and weighted to obtain a discriminant features. The network outputs are then used for classification purpose by exploiting the SVM technique. The conducted experiments showed that QPCANet outperforms RGB PCANet and Gray PCANet in color image classification. Additionally, we experimentally verified that QPCANet performs well in color images databases, which contains many rotation type images. In this work, we have shown that QPCANet is a powerful tool for color image classification and, when compared to PCANet, it allows deriving features that are less sensitive to rotation.

**Acknowledgement**

This work was supported by the National Basic Research Program of China under Grant 2011CB707904, by the NSFC under Grants 61201344, 61271312, 11301074, and by the SRFDP under Grants 20110092110023 and 20120092120036, the Project-sponsored by SRF for ROCS, SEM, and by Natural Science Foundation of Jiangsu Province under Grant BK2012329 and by Qing Lan Project. This work was also supported by INSERM postdoctoral fellowship.

**Pictures and Tables**

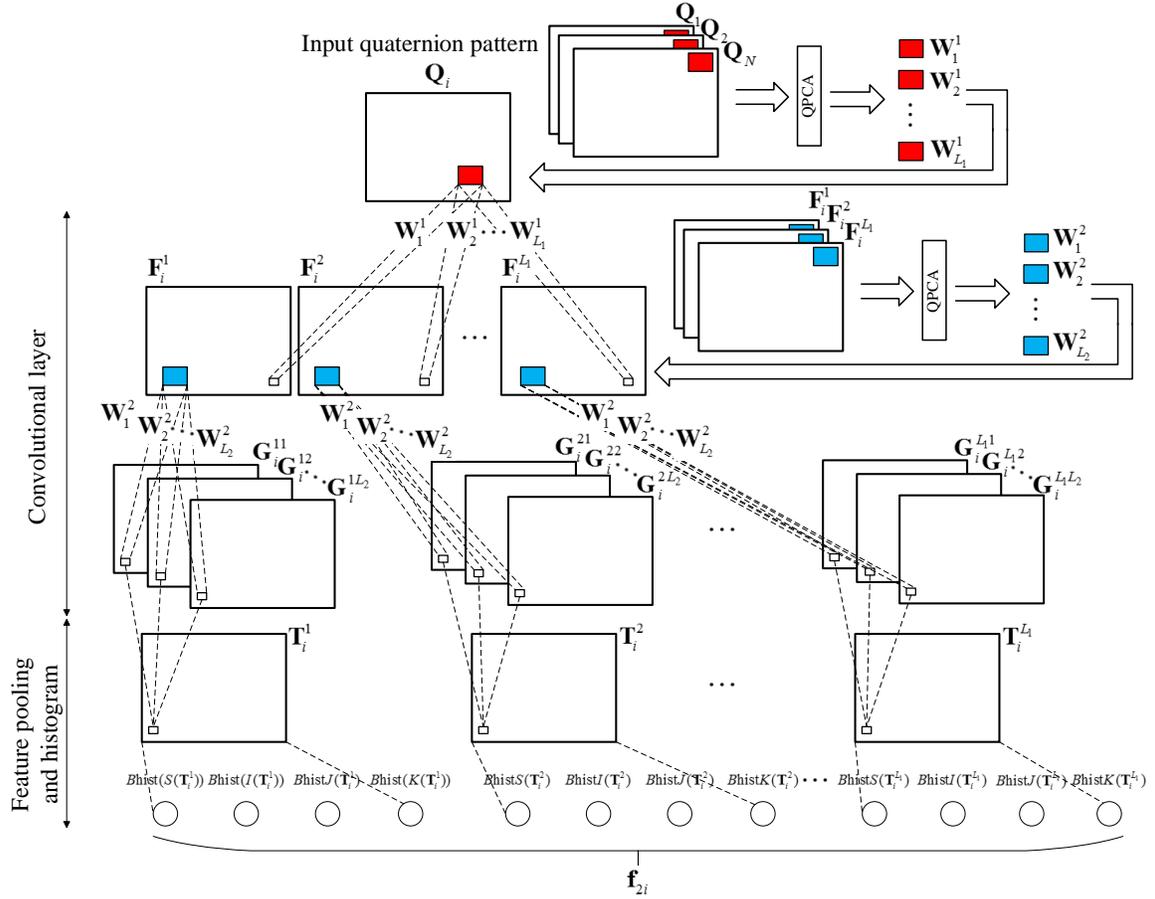

**Fig. 1.** Architecture of two-stage QPCANet

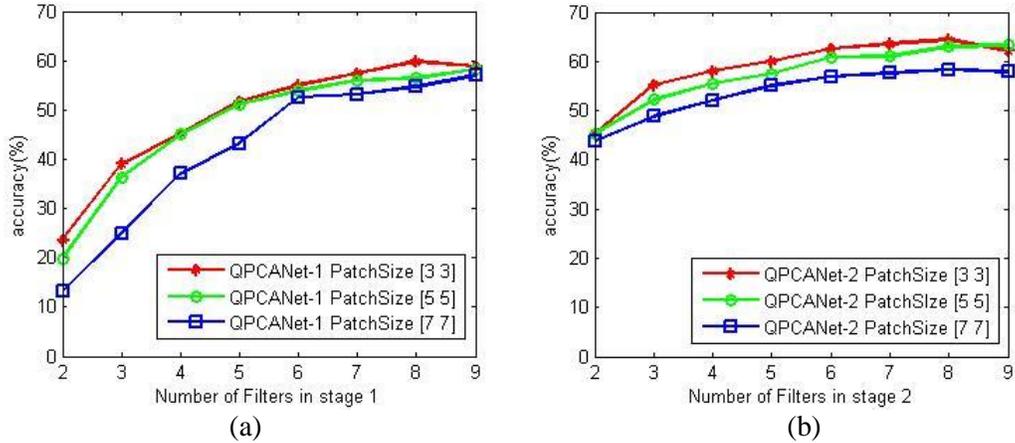

**Fig. 2.** Recognition accuracy of QPCANet on Caltech-101 database for different number of QPCA filters and by changing patch size from $3 \times 3$ to $7 \times 7$. (a) The number of QPCA filters varies from 2 to 9 in one stage. (b) The number of QPCA filters varies from 2 to 9 in the second stage with $L_1$=8

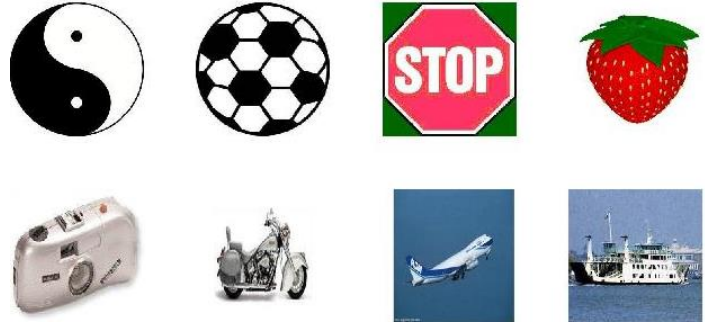

**Fig. 3.** Some pictures have the same main colors but are not belonging to the same category of objects

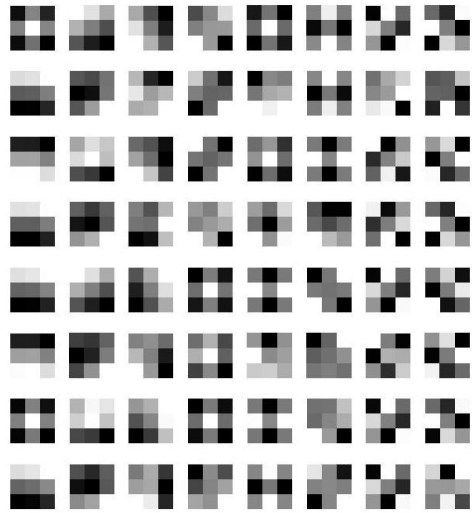

**Fig. 4.** QPCA filters in the first stage (rows 1 to 4) and second stage (rows 5 to 8)

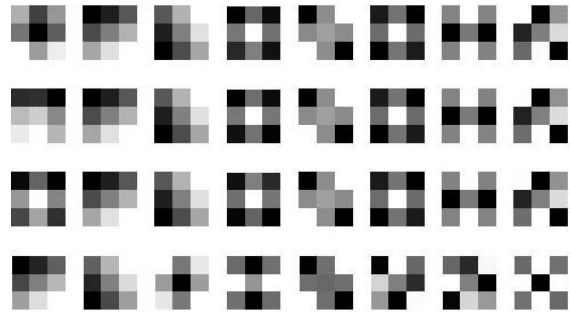

**Fig. 5.** RGB PCA filters in the first stage and second stage

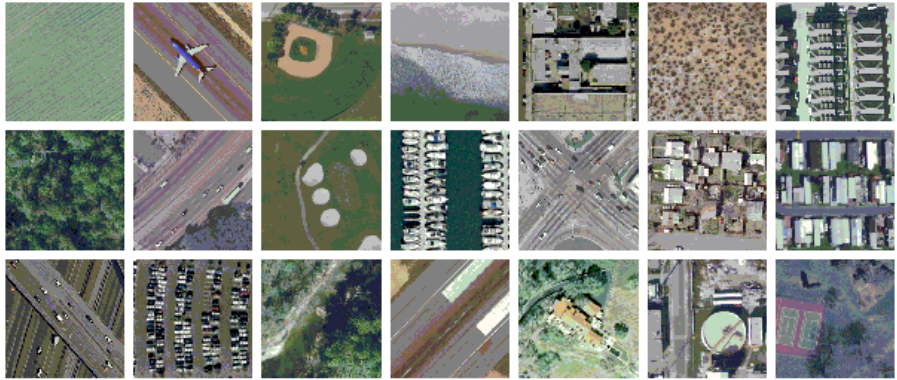

**Fig. 6.** All representative examples of classes in the UC Merced land use database

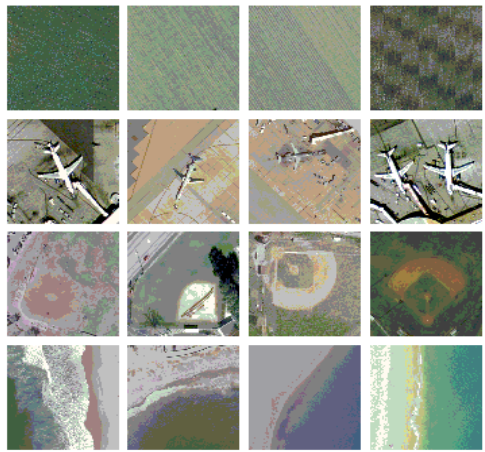

**Fig. 7.** Rotation information in UC Merced land use database

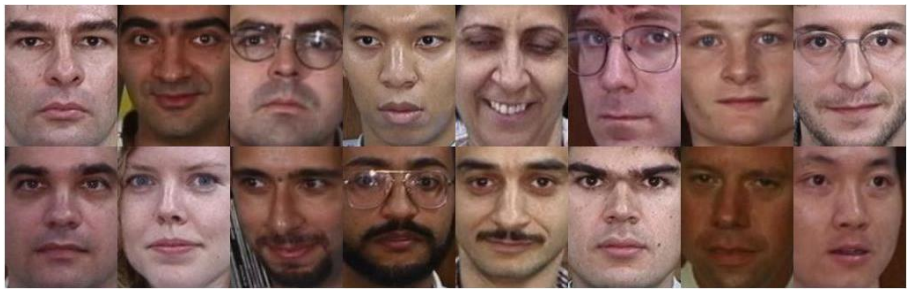

**Fig. 8.** Sixteen individuals in Georgia Tech face database

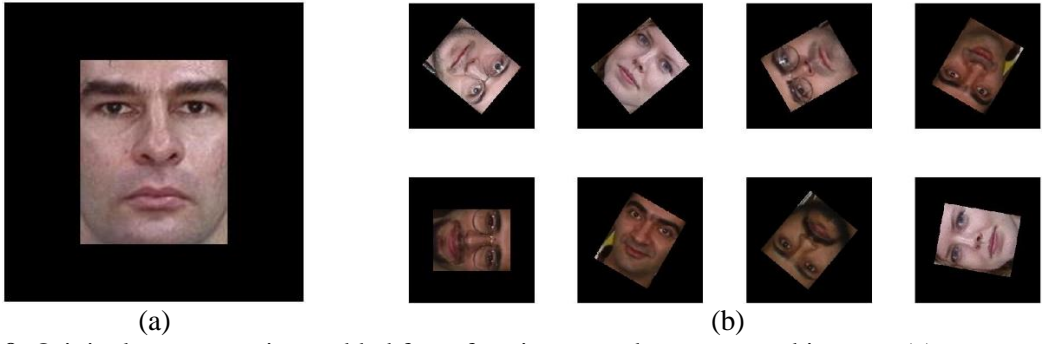

| (a) | (b) |

**Fig. 9.** Original representative padded front face image and some rotated images. (a) Representative padded front face image. (b) Some rotated face images in the new database.

**Table 1**
The accuracy (%) of color image classification in Caltech-101 database

| Patch Size | QPCANet-1 | QPCANet-2 | RGB PCANet-1 | RGB PCANet-2 | Gray PCANet-1 | Gray PCANet-2 |
|---|---|---|---|---|---|---|
| 3 ×3 | 58.59 | **63.78** | 50.33 | 63.50 | 55.69 | 59.90 |
| 5 ×5 | 57.55 | 60.28 | 47.27 | **63.31** | 55.43 | 60.09 |
| 7 ×7 | 56.12 | 57.40 | 47.08 | **59.38** | 54.06 | 54.58 |

**Table 2**
Recognition rate (%) of land use images on UC Merced land use database

| Patch Size | QPCANet-1 | QPCANet-2 | RGB PCANet-1 | RGB PCANet-2 | Gray PCANet-1 | Gray PCANet-2 |
|---|---|---|---|---|---|---|
| 3 ×3 | 70.24 | **79.05** | 65.24 | 73.33 | 65.00 | 66.67 |
| 5 ×5 | 68.81 | **70.71** | 64.29 | 67.14 | 57.14 | 58.81 |
| 7 ×7 | **68.10** | 67.62 | 62.14 | 63.10 | 55.24 | 56.90 |

;
Face recognition rates (%) of different networks on Georgia Tech face database

| Patch Size | QPCANet-1 | RGB PCANet-1 | Gray PCANet-1 |
|---|---|---|---|
| 3 ×3 | **100** | 99.2 | 99.2 |
| 5 ×5 | **100** | 98.8 | 99.2 |
| 7 ×7 | **100** | 99.6 | 99.2 |

**Table 4**
Comparison of face recognition rates (%) of rotation on Georgia Tech face database

| Patch Size | QPCANet-1 | QPCANet-2 | RGB PCANet-1 | RGB PCANet-2 | Gray PCANet-1 | Gray PCANet-2 |
|---|---|---|---|---|---|---|
| 3 ×3 | **82.39** | 77.12 | 80.56 | 74.17 | 60.22 | 47.77 |
| 5 ×5 | **84.94** | 79.17 | 71.49 | 68.61 | 69.44 | 61.66 |
| 7 ×7 | **79.77** | 76.94 | 70.83 | 65.56 | 70.56 | 63.61 |

**Table 5**
The accuracy (%) of texture classification on CURet dataset

| Patch Size | QPCANet-1 | QPCANet-2 | RGB PCANet-1 | RGB PCANet-2 | Gray PCANet-1 | Gray PCANet-2 |
|---|---|---|---|---|---|---|
| 3 ×3 | 96.94 | **98.66** | 92.28 | 96.36 | 90.71 | 93.70 |
| 5 ×5 | 97.81 | **98.57** | 95.60 | 98.08 | 95.60 | 96.56 |
| 7 ×7 | 98.05 | **98.40** | 96.29 | 98.03 | 96.03 | 96.20 |